\documentclass[letterpaper,journal]{IEEEtran}
\usepackage{amsmath,amssymb,amsfonts,amsthm}
\usepackage{algorithmic}
\usepackage{algorithm}
\usepackage{array}
\usepackage[caption=false,font=normalsize,labelfont=sf,textfont=sf]{subfig}
\usepackage{textcomp}
\usepackage{stfloats}
\usepackage{url}
\usepackage{verbatim}
\usepackage{graphicx}
\usepackage{tikz}
\usetikzlibrary{arrows.meta}
\usepackage{pgfplots}
\pgfplotsset{compat=1.18}
\definecolor{qwenblue}{HTML}{1F77B4}
\definecolor{gemmared}{HTML}{D62728}
\definecolor{qwengreen}{HTML}{2CA02C}
\usepackage{float}
\usepackage{cite}
\usepackage{hyperref}
\hypersetup{hidelinks}
\usepackage{fbox}
\usepackage{tabularx}
\usepackage{multirow}

\newcolumntype{F}[1]{%
  >{\hsize=#1\hsize\raggedright\arraybackslash\scriptsize}X%
}

\newcolumntype{Y}[1]{%
  >{\hsize=#1\hsize\raggedright\arraybackslash}X%
}

\newcolumntype{R}[1]{%
  >{\hsize=#1\hsize\raggedleft\arraybackslash}X%
}

\usepackage{glossaries}
\newacronym{atc}{ATC}{air traffic control}
\newacronym{icao}{ICAO}{International Civil Aviation Organization}
\newacronym{ltlf}{LTL$_f$}{linear temporal logic on finite traces}
\newacronym{tcas}{TCAS}{traffic alert and collision avoidance system}
\newacronym{ra}{RA}{resolution advisory}
\newacronym{adsb}{ADS-B}{automatic dependent surveillance-broadcast}
\newacronym{fl}{FL}{flight level}
\newacronym{ltl}{LTL}{linear temporal logic}
\newacronym{llm}{LLM}{Large Language Model}
\newacronym{mtl}{MTL}{Metric temporal logic}
\newacronym{asr}{ASR}{Automatic Speech Recognition}
\newacronym{rv}{RV}{runtime verification}
\glsdisablehyper

\newcounter{example}
\newenvironment{example}[1][]{\refstepcounter{example}\par\medskip\noindent
   \textbf{Example~\theexample. #1} \rmfamily}{\qed\medskip}

\newcommand{\atom}[1]{\text{\small\texttt{#1}}}

\newcommand\reqGrounding[0]{$\textbf{R}_1$}
\newcommand\reqIntegration[0]{$\textbf{R}_2$}
\newcommand\reqDeadlines[0]{$\textbf{R}_3$}
\newcommand\reqVerdicts[0]{$\textbf{R}_4$}

\newcommand{\hbF}{\tikz[baseline=-0.5ex]\fill (0,0) circle (0.6ex);}
\newcommand{\hbE}{\tikz[baseline=-0.5ex]\draw (0,0) circle (0.6ex);}
\newcommand{\hbH}{\tikz[baseline=-0.5ex]{\draw (0,0) circle (0.6ex);\fill (0,-0.6ex) arc (-90:90:0.6ex) -- cycle;}}

\begin{document}
\sloppy

\title{
Formal, Executable and Explainable Runtime Monitoring of Spoken Air Traffic Control Operational Procedures
}


\author{
  Roberto~Luvini, Giacomo~Longo, Alessandro~Armando, and Enrico~Russo

\IEEEcompsocitemizethanks{\IEEEcompsocthanksitem 
This work has been submitted to the IEEE for possible publication. Copyright may be transferred without notice, after which this version may no longer be accessible. Manuscript received ...; R. Luvini (Corresponding author) and A. Armando are with the Department of Informatics, Bioengineering, Robotics, and Systems Engineering (DIBRIS), University of Genoa, Italy.
Email: S4979483@studenti.unige.it; alessandro.armando@unige.it. G. Longo and E. Russo are with CASD - University School of Advanced Defense Studies, Rome, Italy. Email: \{name.surname\}@unicasd.it.\protect

}
}


\markboth{
    IEEE Transactions on Intelligent Transportation Systems
}
{}


\maketitle

\begin{abstract}
Air traffic control procedures are executed through spoken exchanges between controllers and pilots.
These interactions are essential to the safety of air transportation: failures in their execution can create severe operational hazards, as evidenced by past fatal accidents.
Assessing whether an instruction has been followed requires relating what was said to the aircraft concerned, its state, and the obligations that pilots must meet.
We present a runtime verification framework that monitors such procedures by checking controller-pilot exchanges, surveillance data, and onboard observations.
The framework parses radio communications into events linked to the entities they concern and merges them with surveillance and onboard observations into a time-stamped trace.
The ICAO-derived obligations as formalized as temporal formulas with explicit time bounds and evaluated over execution traces.
Every violation is reported along with the breached obligations and the observations that support the verdict.
With real traffic, the complete pipeline reaches an F1 of 0.85 against blind human-annotated violations; in 1,495 synthetic situations derived from two public corpora, the monitor logic returns the expected verdict in every case.
In two historical accidents reconstructed from official investigation reports, the monitor identifies the same procedural deviations documented by the investigators.
\end{abstract}

\begin{IEEEkeywords} Air traffic control, Runtime verification, Formal methods, Methods for safety, Air transportation.
\end{IEEEkeywords}

\section{Introduction}\label{sec:intro}

\IEEEPARstart{T}{ransportation} procedures coordinate vehicle operators and control centers. 
In aviation, these roles correspond to pilots and \gls{atc}: controllers issue clearances and instructions, pilots acknowledge and execute them, and both rely on standardized phraseology. 
These exchanges create obligations that must be interpreted, tracked, and verified as the situation evolves.

For \gls{atc}, determining whether an instruction has been followed cannot rely on the transcript alone.
It requires interpreting what was said, by whom, to whom it was addressed, whether it was acknowledged, and whether the aircraft state evolved accordingly.
This involves combining speech content, roles, aircraft identity, system state, and timing, as well as reconstructing the obligations created by the radio exchange and verifying them against operational evidence.
When a deviation occurs, the controller must be able to identify which obligation failed and which observations support that conclusion.
Consequently, the safety-critical activity of determining whether ongoing pilot-controller interactions satisfy all applicable obligations ultimately relies on human operators.

In this paper, we make this reasoning explicit through four monitoring requirements: \emph{observation grounding} (\reqGrounding), which links each utterance to the aircraft and operational entity it concerns, such as a runway or vertical clearance; \emph{multi-source integration} (\reqIntegration), which relates the radio exchange to surveillance and onboard observations; \emph{temporal and timeliness reasoning} (\reqDeadlines), which tracks and relates acknowledgments, active obligations, deadlines, and superseded clearances; and \emph{evidence-backed verdicts} (\reqVerdicts), which connect each deviation to the corresponding failed obligation and observations.

These requirements reflect routine checks in \gls{atc}, but meeting them continuously across multiple aircraft and radio exchanges under stringent real-time constraints is increasingly challenging as traffic grows and staffing often remains constrained. 
Worldwide scheduled operations reached 37 million departures in 2024~\cite{icao_sr2025}.
In Europe, \gls{atc} capacity and staffing accounted for more than half of the 22.4 million minutes of en-route delay in 2024, the worst level since 2001~\cite{eurocontrol_prr2024}. 
Such workload pressures occur in a safety-critical domain, where runway collisions remain a major risk~\cite{easa_asr_2025} and procedural deviations have contributed to documented accidents, including the cases examined in ~\S\ref{sec:casestudies}.

Automated assistance could help address this burden, but only if spoken exchanges can be checked against the evolving operational state.
Advances in speech transcription and \gls{llm} based semantic parsing now make real-time extraction of operational events from radio communications feasible and automated support for spoken \gls{atc} procedure monitoring an active research direction.
However existing approaches cover the above requirements only in part.

We close this gap with a novel runtime verification framework that turns spoken radio exchanges into entity-linked operational events, combines them with surveillance and onboard state in a formal model of the procedure, and evaluates the resulting trace to produce time- and evidence-linked verdicts.
Encoding the procedures as temporal formulas gives them a mathematically precise semantics and makes compliance machine-checkable.

This paper presents the following key contributions:
\begin{itemize}
 \item A temporal formalism for encoding spoken operational procedures as formal specifications with metric time bounds and precedence among obligations; to the best of our knowledge, the first temporal-logic formalization of controller and pilot obligations. 
 \item An automated trace-construction pipeline that transforms spoken radio exchanges and heterogeneous operational data into unified execution traces.
 \item A verification framework that instantiates and evaluates these specifications over operational traces, producing verdicts linked to the violated obligation and supporting observations.
 \item An empirical evaluation on real traffic and synthetic situations, showing that the framework applies to genuine controller-pilot communications and, in particular, detects rare safety-critical configurations under controlled conditions.
 \item A validation of the framework on two documented accidents, \"Uberlingen (2002) and Comair~5191 (2006), reconstructed from the official investigation reports, showing that it identifies the reported procedural deviations with a time margin before each impact.
\end{itemize}
The framework supports three operational uses: $(i)$~\emph{real-time assistance}, running alongside operations to flag anomalies as they arise; $(ii)$~\emph{post-operation review}, auditing recorded traces for safety assessments; and $(iii)$~\emph{controller training}, providing active feedback on procedural compliance.
Its deployment as an operational technology could significantly improve aviation safety by reducing the risk that procedural deviations go undetected, a recurring contributing factor in serious accidents.
\par \emph{Structure of the paper.}
The rest of the paper is organized as follows.
In ~\S\ref{sec:background}, we provide background on \gls{atc} procedures and on temporal-logic specifications over finite traces.
~\S\ref{sec:casestudies} presents two motivating examples for the requirements.
In ~\S\ref{sec:formal-model}, we introduce our formal model for procedural trace verification and, in ~\S\ref{sec:framework}, the architecture that implements it.
We evaluate the framework in ~\S\ref{sec:evaluation}, discuss the results in ~\S\ref{sec:discussion}, compare with the related work in ~\S\ref{sec:related}, and conclude the paper in ~\S\ref{sec:conclusion}.

\section{Background}\label{sec:background}

\subsection{\texorpdfstring{\Acrlong{atc}}{ATC} procedures and monitoring sources}
\label{sec:atcproc}
In \gls{atc}, a controller is responsible for preventing collisions between aircraft and for maintaining an orderly flow of traffic~\cite[\S2.2,~\S2.3.1]{InternationalCivilAviationOrganization2018Annex11}.
The controller and the pilots communicate by voice over a radio channel, exchanging spoken messages in real time as each flight progresses. 
On this channel, each aircraft is addressed by its radio identifier, namely the \emph{call sign}, typically an operator designator followed by a flight number~\cite{InternationalCivilAviationOrganization2016Annex10V2}
, e.g., \texttt{AZA545}.
These exchanges are not free-form.
ICAO Doc~4444~\cite{InternationalCivilAviationOrganization2016Procedures} codifies both the procedure and the phraseology used to express clearances, instructions, and acknowledgments~\cite[Ch.~12]{InternationalCivilAviationOrganization2016Procedures}. 
A \emph{clearance} authorizes an aircraft to proceed under conditions specified by the controller, whereas an \emph{instruction} directs the pilot to perform a specific action~\cite[Ch.~1]{InternationalCivilAviationOrganization2016Procedures}, such as adopting a particular heading, speed, or assigned level.
Here, a \emph{level} denotes the aircraft's assigned vertical position and is stated as a \gls{fl}, referenced to a standard pressure datum~\cite[Ch.~1]{InternationalCivilAviationOrganization2016Procedures}, e.g., \texttt{FL120}.
The pilot acknowledges the safety-related parts of these messages by reading them back to the controller, who listens and corrects any discrepancy the readback reveals~\cite[\S4.5.7.5]{InternationalCivilAviationOrganization2016Procedures}.

Instruction compliance cannot be assessed from the radio exchange alone. It requires information on the aircraft state, e.g., position and speed, obtained either by external observation, as in radar surveillance, or automatically transmitted by \gls{adsb}~\cite[Ch.~1]{InternationalCivilAviationOrganization2014Annex10V4}.
Thus, the radio exchange specifies the intended action, whereas surveillance provides the observed behavior of the aircraft.
Compliance is the comparison of the two.

Operational monitoring today relies on surveillance-derived safety nets: short-term conflict alert warns the controller of a predicted infringement of separation minima within its warning time~\cite{eurocontrol2007stca}, and, on the airport surface, advanced surface movement guidance and control systems alert the controller to runway incursions~\cite{InternationalCivilAviationOrganization2004Asmgcs}.

Aircraft behavior may also be constrained by onboard safety systems.
The \gls{tcas} is a transponder-based airborne collision-avoidance system defined by \gls{icao}~\cite{InternationalCivilAviationOrganization2012Acas}. 
It tracks nearby aircraft and, independently of \gls{atc}, issues a \gls{ra} instructing the pilot to climb or descend when a collision risk becomes imminent.

\subsection{Temporal-logic specifications over finite traces}
\label{sec:ltl}
\Gls{ltl} is a formal language for describing properties of how a system behaves over time, and for checking whether a given behavior satisfies them~\cite{pnueli1977temporal}. Its basic facts are \emph{atomic propositions} (atoms, for short): elementary statements that are either true or false at a single point of the behavior. A \emph{temporal formula} combines atomic propositions through the boolean connectives and a set of \emph{temporal operators} that relate points of the behavior across time. The standard operators are \emph{globally} (\(\mathbf{G}\)), a property holding at the current point and every later one; \emph{next} (\(\mathbf{X}\)), a property holding at the immediately following point; \emph{until} (\(\mathbf{U}\)), one property holding until a second occurs; \emph{weak until} (\(\mathbf{W}\)), the same without requiring the second to occur; and \emph{eventually} (\(\mathbf{F}\)), a property holding at the current point or some later one.

A formula is evaluated over a \emph{trace}, a sequence of observations produced by the behavior. 
Each observation records which atomic propositions hold at that point. 
Formally, this evaluation is defined by a \emph{satisfaction relation}, which states whether a formula holds at a given \emph{position} of the trace. 
Temporal operators evaluate the sequence of observations based on their relative position, establishing the constraints imposed by a temporal property.
Classical \gls{ltl} is interpreted over infinite traces, whereas \gls{ltlf}~\cite{degiacomo2013ltlf} evaluates temporal specifications over finite traces, where the sequence of observations has a last position.

The temporal operators of \gls{ltl} and \gls{ltlf} constrain observations occurrence and ordering, but not the elapsed time between them. \Gls{mtl} extends temporal logic with explicit time bounds on temporal requirements~\cite{koymans1990mtl}. \Gls{rv} checks an observed behavior against a temporal-logic specification~\cite{bauer2011}, with monitors for data values, finite traces, and onboard use~\cite{basin2015monpoly,kallwies2023anticipatory,r2u2,moosbrugger2017r2u2}.



\section{Case Studies}
\label{sec:casestudies}

\subsection{Selection and operational coverage}

We use two well-documented accidents as validation scenarios. \emph{\"Uberlingen (2002)} involved two aircraft, at the same level on crossing tracks; all 71 people on board died~\cite{bfu2004ueberlingen}. In \emph{Comair~5191 (2006)}, an aircraft cleared to taxi to runway~22 at Lexington instead took off from the shorter, unlit runway~26 and overran it; 49 of the 50 people on board died~\cite{ntsb2007comair5191}.

We selected these two cases because they are complementary and well documented: they cover airborne and surface operations, combine radio, surveillance, and onboard observations in different proportions, and both stem from failures to meet procedural obligations.
Each was reconstructed in detail by an authoritative investigation carried out by Germany's BFU and the U.S.\ NTSB, respectively.

\subsection{Monitoring requirements exposed by the cases}

The official reports first show the need for evidence-backed verdicts (\reqVerdicts): as post-hoc reconstructions, they link each procedural deviation to the governing obligation, aircraft, instant, and recorded observations. 
The analysis below shows why detecting these deviations requires \reqGrounding--\reqDeadlines.
 
\noindent
\textbf{\"Uberlingen}.
The deviation is invisible in the transcript.
It appears only when the controller's descent instruction and the collision-avoidance advisories are associated with the instructed aircraft (\reqGrounding) and related to its observed vertical motion (\reqIntegration).
During the final minute, an advisory commanded a climb while the controller's standing instruction required descent~\cite{bfu2004ueberlingen}. Under ICAO collision-avoidance procedures, an active advisory takes precedence over controller instructions ~\cite[\S12.3.1.2]{InternationalCivilAviationOrganization2016Procedures}~\cite[\S5.2.1.14]{InternationalCivilAviationOrganization2012Acas}.
So which obligation governed depends on whether each instruction was issued before or after the advisory became active (\reqDeadlines).

\noindent
\textbf{Comair~5191}.
The deviation becomes apparent only after associating the taxi clearance and its assigned runway with the correct aircraft (\reqGrounding), and comparing that authorization with the runway actually occupied (\reqIntegration).
The procedural failure also depends on event order: the aircraft entered runway~26 and began its takeoff roll without an authorization for that runway. 
Detecting the deviation requires tracking whether the relevant authorization existed before runway entry and takeoff (\reqDeadlines).

\section{Formal Model for Procedural Trace Verification}
\label{sec:formal-model}

The procedures considered in this work involve interactions between air traffic controllers, pilots, aircraft states, and safety-critical onboard systems.
At the operational level, they refer to domain-specific facts such as issued clearances, pilot readbacks, assigned flight levels, and runway authorizations, codified by the \gls{icao} in Doc~4444, together with observed aircraft positions and \glspl{ra} issued by the \gls{tcas} (see ~\S\ref{sec:atcproc}).
For automated verification, these aeronautical concepts must be represented in a uniform, mathematically precise form.

To this end, we distinguish between the operational vocabulary and its formal abstraction.
The operational vocabulary identifies the safety-relevant concepts that may appear in \gls{atc} procedures.
The formal abstraction represents these concepts using standard temporal-logic entities: atomic propositions, finite timed traces, and temporal formulas (see ~\S\ref{sec:ltl}).

In our setting, atomic propositions are parameterized by aeronautical entities (e.g., aircraft identifiers, flight levels, and runways), timed traces collect the propositions that hold at each time instant, and temporal formulas encode the monitored procedural requirements.
Moreover, the observed behavior is finite, since a flight, radio exchange, or surveillance recording eventually ends. 
We therefore adopt the finite-trace semantics of \gls{ltlf}. 
Because procedural requirements may also impose deadlines, we attach timestamps to observations and bound temporal operators by elapsed time, in the manner of \gls{mtl}.
This trace-based view connects the operational vocabulary to temporal formulas for procedural compliance.
Based on this abstraction, we define the formal model used for procedural trace verification as
\[
\mathcal{M} =
\langle
\mathcal{AP},
\Pi,
\Phi,
\models
\rangle
\]
where \(\mathcal{AP}\) is a set of atomic propositions; \(\Pi\) is the set of finite timed traces; \(\Phi\) is the finite-trace temporal language used to encode procedural requirements; and \(\models\) is the satisfaction relation used to evaluate formulas at positions.

The following subsections detail each component.

\subsection{Atomic propositions}
\label{sec:atomvocabulary}


The set \(\mathcal{AP}\) collects the facts that can be observed and used in the temporal formulas.
Its propositions are the ground atomic formulas of a many-sorted first-order vocabulary.
The sorts are the kinds of aeronautical entities that the procedures refer to, such as aircraft, flight levels, headings, speeds, and runways, together with a sort \(\mathit{Agent}\) for the speakers and a sort \(\mathit{Utterance}\) for the spoken content; for a given recording, each entity sort comes with the finite set of constants naming the observed values, such as the call signs heard on the radio.

Facts asserted by a person are built around the predicate \(\mathit{Says}\), of sort \(\mathit{Agent} \times \mathit{Utterance}\): the constants \atom{atc} and \atom{pilot} of sort \(\mathit{Agent}\) denote the two speaking roles, and each spoken entry of the operational vocabulary contributes a function symbol that builds the uttered content, such as \(\atom{cmd\_descend}\) of sort \(\mathit{Aircraft} \times \mathit{Level} \rightarrow \mathit{Utterance}\).
Facts established during trace construction, whether observed by surveillance or computed, are expressed by predicates over the same sorts, such as \(\atom{descending}\) over \(\mathit{Aircraft}\).
Computed facts are obtained from the observed ones, spoken or surveillance-derived, through fixed relations; these relations, together with the vocabulary, define the framework's domain ontology. 

A controller descent instruction, for instance, is a proposition of the form \(\mathit{Says}\bigl(\atom{atc},\allowbreak\ \atom{cmd\_descend}(a,\ell)\bigr)\), where \(a\) and \(\ell\) stand for constants of sorts \(\mathit{Aircraft}\) and \(\mathit{Level}\): they are variables of the metalanguage, not of the language, which is ground.
Propositions are grouped by procedural area, such as descent, approach, ground movement, or emergency handling.
Since the vocabulary has finitely many symbols and each sort finitely many constants, there are finitely many ground atomic formulas, and \(\mathcal{AP}\) collects all of them.

We use four families of propositions:
$(i)$ \emph{speech propositions} represent controller and pilot utterances, including clearances, instructions, acknowledgments, and readbacks;
$(ii)$ \emph{air-state propositions} represent surveillance-derived aircraft state, such as level, position, and vertical motion;
$(iii)$ \emph{surface-state propositions} represent runway and taxiway occupancy;
$(iv)$ \emph{derived propositions} represent facts computed from the observed ones, such as whether a readback matches the corresponding instruction or an authorization remains active across observations, together with the resolution advisories we reconstruct from the aircraft trajectories or from the radio.
Some propositions are inertial: trace construction records them as holding until a later observation ends them; others hold only at the observations that report them.

\begin{example}
\label{ex:atomvoc}
A descent instruction is a controller utterance directing an aircraft to descend to an assigned flight level, with the required readback as described in ~\S\ref{sec:atcproc}.
The corresponding instruction, readback, and aircraft-state facts have the forms
\[
\begin{aligned}
&\mathit{Says}\bigl(\atom{atc},\ \atom{cmd\_descend}(a,\ell)\bigr),\\
&\mathit{Says}\bigl(\atom{pilot},\ \atom{readback\_descend}(a,\ell)\bigr),\\
&\atom{descending}(a)
\end{aligned}
\]

All three belong to the descent area. The spoken forms share \(a\) and \(\ell\), i.e., the same addressed aircraft and assigned level, and differ in the speaking role and in the utterance built, one carrying the controller's instruction and the other the pilot's readback. The state form mentions only the aircraft \(a\), since it records the observed vertical motion of that aircraft; no agent asserts it, as it is obtained from surveillance.
\end{example}


\subsection{Timed traces}
\label{sec:timedtraces}

A \emph{timed trace} \(\pi \in \Pi\) is a finite sequence of observations
\[
\pi = (t_1, L_1),(t_2, L_2),\dots,(t_n, L_n)
\]
where \(n \ge 1\) is the length of the trace.
Each pair \((t_i,L_i)\) is the observation at position \(i\): \(t_i \in \mathbb{N}\) is the instant in seconds from the start of the recording, and \(L_i \subseteq \mathcal{AP}\) is the set of propositions that hold at that instant.

Explicit timestamps are needed to check procedural deadlines stated in seconds: observations arrive at irregular intervals, so order alone does not determine elapsed time.
They are monotone, \(t_i \le t_{i+1}\), so that observations follow their recorded order. 
Equal timestamps are admitted, since independent sources may report at the same instant.

\begin{example}
\label{ex:timedtrace}
Consider the descent facts from Example~\ref{ex:atomvoc}.
Suppose the controller instructs \atom{AZA545} to descend to \atom{FL120}, the crew reads the instruction back eight seconds later, and surveillance then reports the aircraft descending. 
These observations form the trace $\pi$
{\small
\begin{align*}
&\bigl(0,\ \{\mathit{Says}(\atom{atc},\ \atom{cmd\_descend}(\atom{AZA545},\atom{FL120}))\}\bigr),\\
&\bigl(8,\ \{\mathit{Says}(\atom{pilot},\ \atom{readback\_descend}(\atom{AZA545},\atom{FL120}))\}\bigr),\\
&\bigl(20,\ \{\atom{descending}(\atom{AZA545})\}\bigr)
\end{align*}
}
The controller's instruction holds at position~1, at \(t_1 = 0\). 
Its readback holds at position~2, at \(t_2 = 8\). 
The descent reported by surveillance, recorded as the air-state proposition \(\atom{descending}(\atom{AZA545})\), holds at position~3, at \(t_3 = 20\). 
Thus, the trace records both the order of the three facts and their unequal elapsed times, all measured in seconds from the start of the recording.
\end{example}

\subsection{Temporal-logic patterns and \texorpdfstring{\gls{atc}}{ATC} formula families}
\label{sec:temporalpatterns}

A temporal formula \(\varphi \in \Phi\) is built from propositions \(p \in \mathcal{AP}\), boolean connectives \(\neg, \wedge, \vee, \rightarrow\), and temporal operators over trace positions (see ~\S\ref{sec:ltl}). 

The formulas follow five recurring logical patterns, each capturing a common procedural dependency such as response, compliance, precedence, or exception handling.
In the patterns, \(\alpha\) and \(\beta\) stand for propositional formulas: boolean combinations of propositions in \(\mathcal{AP}\), with no temporal operators.

Several of these patterns use an outer \(\mathbf{G}\) to make the requirement apply at every trace position. 

A \emph{bounded-response} pattern has the form
\[
\mathbf{G}\bigl(\alpha \rightarrow \mathbf{F}_{\le\delta}\,\beta\bigr)
\]
requiring every trigger \(\alpha\) to be followed by a position satisfying
\(\beta\) within \(\delta\) seconds. The metric operator
\(\mathbf{F}_{\le\delta}\) is the only operator that uses timestamps. 
The other temporal operators reason over the order of trace positions. 

An \emph{invariant pattern} has the form 
\[
\mathbf{G}(\alpha \rightarrow \beta)
\]
requiring every position that satisfies \(\alpha\) to satisfy \(\beta\) as well. 
A \emph{precedence pattern} has the form
\[
\neg\,\alpha\ \mathbf{W}\ \beta
\]
requiring that no action \(\alpha\) occur before an authorization
\(\beta\). 
Weak until is used because a trace in which the action never occurs should not be reported as a violation merely because no authorization
appears.

A \emph{pending-response} pattern has the form
\[
\mathbf{G}\bigl(\alpha \rightarrow
\neg\mathbf{X}\neg(\neg\,\alpha\ \mathbf{W}\ \beta)\bigr)
\]
requiring that, after a trigger \(\alpha\), no further trigger of the same kind
appear before the response \(\beta\). 
The guard \(\neg\mathbf{X}\neg\) is weak next: it coincides with strong next except at the final trace position, where it avoids reporting a violation solely because no next position exists.

Finally, a \emph{bounded-prohibition} pattern has the form
\[
\mathbf{G}\bigl(\alpha \rightarrow
\neg\,\mathbf{F}_{\le\delta}\beta\bigr)
\]
requiring that, whenever \(\alpha\) holds, no position satisfying \(\beta\) occur within \(\delta\) seconds.

A pattern is not itself a formula of \(\Phi\): it denotes the set of formulas of that form.
With \(\alpha\) and \(\beta\) so restricted, no formula of one pattern belongs to another.
The formulas of one pattern differ only in the propositional formulas for \(\alpha\) and \(\beta\) and, in the bounded patterns, in the time bound \(\delta\); a pattern thus mentions no specific aircraft, value, or utterance.

Specializing these patterns to the \gls{atc} domain gives the families of formulas below.
The set is extensible: new procedures are codified as further instances of the same five patterns.



Each \gls{atc} family selects, within one of the patterns above, the form of each propositional formula and, where required, the operational time bound.
A variable of the metalanguage, such as \(a\) or \(\ell\), occurring in more than one form of the same family denotes the same constant in each.
The normative sources fix the obligations but not their deadlines, except for the five-second resolution-advisory response~\cite[\S4.1.4.2]{InternationalCivilAviationOrganization2012Acas}.
The remaining bounds, such as the thirty-second readback window, are operational choices; pilot response-time statistics~\cite{lutz2022response} provide the empirical basis for selecting them.
Table~\ref{tab:families} summarizes the families considered in this work.
For each family, it reports the pattern, its formula, representative forms of \(\alpha\) and \(\beta\), and the normative source from which the procedural requirement is derived.

\begin{table*}[t]
\centering
\footnotesize
\setlength{\tabcolsep}{2pt}
\renewcommand{\arraystretch}{0.95}
\renewcommand{\atom}[1]{\text{\scriptsize\texttt{#1}}}
\caption{Families with pattern, formula, representative forms of \(\alpha\) and \(\beta\), and source.}
\label{tab:families}
\renewcommand\tabularxcolumn[1]{m{#1}}
\begin{tabularx}{\linewidth}{@{}
  Y{0.80}|
  Y{0.95}|
  Y{1.16}|
  Y{1.50}|
  R{0.59}
@{}}
\hline
\multicolumn{1}{c}{\textbf{Family}} & \multicolumn{1}{c}{\textbf{Pattern}} & \multicolumn{1}{c}{\(\boldsymbol{\alpha}\)} & \multicolumn{1}{c}{\(\boldsymbol{\beta}\)} & \multicolumn{1}{c}{\textbf{Source}} \\
\hline\hline
readback
  & \multirow{3}{=}{\centering bounded-response\par\(\mathbf{G}(\alpha \rightarrow \mathbf{F}_{\le\delta}\beta)\)}
  & \(\mathit{Says}(\atom{atc},\allowbreak\ \atom{cmd\_descend}(a,\ell))\)
  & \(\mathit{Says}(\atom{pilot},\allowbreak\ \atom{readback\_descend}(a,\ell))\)
  & \cite[\S4.5.7.5]{InternationalCivilAviationOrganization2016Procedures}\par\cite[\S5.2.1.9.2.1]{InternationalCivilAviationOrganization2016Annex10V2} \\
state-consistency
  &
  & \(\mathit{Says}(\atom{pilot},\allowbreak\ \atom{readback\_descend}(a,\ell))\)
  & \(\atom{descending}(a)\)
  & \cite[\S4.5.7.5.1.1]{InternationalCivilAviationOrganization2016Procedures} \\
\gls{ra}
  &
  & \(\atom{ra\_climb}(a)\)
  & \(\atom{climbing}(a)\)
  & \cite[\S4.1.4.2]{InternationalCivilAviationOrganization2012Acas} \\
\hline
runway-incursion
  & \multirow{2}{=}{\centering invariant\par\(\mathbf{G}(\alpha \rightarrow \beta)\)}
  & \(\atom{on\_runway}(a,r)\)
  & \(\atom{line\_up\_authorized}(a,r)\)
  & \cite[\S7.6.3.1.1.2]{InternationalCivilAviationOrganization2016Procedures}\par\cite[\S7.6.3.1.2.1]{InternationalCivilAviationOrganization2016Procedures} \\
advisory-priority
  &
  & \(\atom{ra\_active}(a)\)
  & \(\neg\,\mathit{Says}(\atom{atc},\allowbreak\ \atom{cmd\_descend}(a,\ell))\)
  & \cite[\S15.7.3.2]{InternationalCivilAviationOrganization2016Procedures} \\
\hline
clearance-precedence
  & \centering precedence\par\(\neg\,\alpha\ \mathbf{W}\ \beta\)
  & \(\atom{landed}(a)\)
  & \(\atom{landing\_cleared}(a)\)
  & \cite[\S7.10.2]{InternationalCivilAviationOrganization2016Procedures}\par\cite[\S6.5.6]{InternationalCivilAviationOrganization2016Procedures} \\
\hline
repeated-instructions
  & \centering pending-response\par\(\mathbf{G}\bigl(\alpha \rightarrow \neg\mathbf{X}\neg(\neg\alpha\,\mathbf{W}\,\beta)\bigr)\)
  & \(\mathit{Says}(\atom{atc},\allowbreak\ \atom{cmd\_descend}(a,\ell))\)
  & \(\mathit{Says}(\atom{pilot},\allowbreak\ \atom{readback\_descend}(a,\ell))\)
  & \cite[\S4.5.7.5]{InternationalCivilAviationOrganization2016Procedures} \\
\hline
advisory-direction
  & \centering bounded-prohibition\par\(\mathbf{G}(\alpha \rightarrow \neg\mathbf{F}_{\le\delta}\beta)\)
  & \(\atom{ra\_climb}(a)\)
  & \(\atom{descending}(a)\)
  & \cite[\S4.1.4.4]{InternationalCivilAviationOrganization2012Acas} \\
\hline
\end{tabularx}
\end{table*}

The bounded-response rows cover three delayed obligations: \emph{readback} links controller instructions to pilot readbacks, \emph{state-consistency} links acknowledged maneuvers to the observed aircraft state, and \emph{resolution-advisory} links advisories to the expected pilot response.

The invariant rows cover authorization and contingency states: \emph{runway-incursion} checks that runway occupancy has a matching authorization, while \emph{advisory-priority} blocks immediate controller instructions during active advisories except for deferrals such as ``when ready''.

The \emph{clearance-precedence} row covers actions that require prior clearance, for example landing or establishing on final approach.
The \emph{repeated-instructions} row prevents a second instruction of the same kind before the first one has been read back.
The \emph{advisory-direction} row covers bounded prohibitions: after a climb advisory, descent is forbidden within the operational window, and vice versa for descend advisories, unless a later advisory reversing the original one, recorded as a derived proposition, is present.

\begin{example}
\label{ex:readbackformula}
For the instruction of Example~\ref{ex:timedtrace}, the \emph{readback} family contains the formula
\[
\begin{aligned}
\mathbf{G}\bigl(\, &\mathit{Says}(\atom{atc},\ \atom{cmd\_descend}(\atom{AZA545},\atom{FL120})) \rightarrow \mathbf{F}_{\le 30}\,\\
&\mathit{Says}(\atom{pilot},\ \atom{readback\_descend}(\atom{AZA545},\atom{FL120}))\,\bigr)
\end{aligned}
\]
The formula is evaluable over the trace of Example~\ref{ex:timedtrace}.
\end{example}

\subsection{Evaluation semantics}
\label{sec:evalsemantics}

The satisfaction relation \(\models\) defines how formulas are evaluated over timed traces. For a trace \(\pi = (t_1, L_1),\dots,(t_n, L_n)\), a position \(i\), and a formula \(\varphi\), the judgment \(\pi,i\models\varphi\) states that \(\varphi\) is satisfied at position \(i\). Each formula is evaluated on the trace, yielding its \emph{verdict}.

Atomic propositions are interpreted by the sets \(L_i\): \(\pi,i\models a\) iff \(a\in L_i\). Boolean connectives have their standard semantics at position \(i\). The temporal operators used in the formulas are defined as follows:
\[
\begin{array}{@{}ll@{}}
\pi,i\models \mathbf{X}\varphi
&\Leftrightarrow i<n \text{ and } \pi,i{+}1\models\varphi,\\[1mm]
\pi,i\models \mathbf{G}\varphi
&\Leftrightarrow \forall j\,(i\le j\le n \Rightarrow \pi,j\models\varphi),\\[1mm]
\pi,i\models \varphi\,\mathbf{U}\,\psi
&\Leftrightarrow \exists k\,(i\le k\le n \land \pi,k\models\psi \\
&\qquad\land\ \forall j\,(i\le j<k \Rightarrow \pi,j\models\varphi)),\\[1mm]
\pi,i\models \varphi\,\mathbf{W}\,\psi
&\Leftrightarrow \pi,i\models \varphi\,\mathbf{U}\,\psi \\
&\qquad\lor\ \forall j\,(i\le j\le n \Rightarrow \pi,j\models\varphi),\\[1mm]
\pi,i\models \mathbf{F}_{\le\delta}\varphi
&\Leftrightarrow \exists k\,(i\le k\le n \land t_k\le t_i+\delta \\
&\qquad\land\ \pi,k\models\varphi).
\end{array}
\]
The operator \(\mathbf{X}\) is strong next: it requires a successor position and is therefore false at the end of the trace. The operator \(\mathbf{G}\) is universal over the suffix starting at \(i\), so a single later counterexample falsifies it. Until, \(\mathbf{U}\), requires a witness position where \(\psi\) holds, with \(\varphi\) holding at every intervening position. Weak until, \(\mathbf{W}\), admits the same witnessed case, but also the case in which \(\psi\) never occurs and \(\varphi\) holds until the trace ends. The bounded finally operator \(\mathbf{F}_{\le\delta}\) is the metric part of the semantics: its witness must occur no later than \(\delta\) seconds after the current timestamp. \(\mathbf{X}\) and \(\mathbf{U}\) enter only through \(\neg\mathbf{X}\neg\) and the definition of \(\mathbf{W}\), respectively.

Every formula is evaluated at the first position of the trace, so its outermost operator ranges over all positions.
We call the procedure that applies \(\models\) to a trace the \emph{monitor}: it runs either offline, over a completed recording, or online, on a still-growing trace, issuing a verdict as new observations arrive.
Offline, over a recording that extends past the deadlines of the obligations triggered in it, the verdict is two-valued: by the end every obligation has been met or missed, so each formula is definitively true or false.

While the trace is still open, a formula may not yet have a definitive verdict: a time-bounded obligation triggered at position \(i\) whose response has not yet been observed is false under \(\models\), since no witness exists yet, although one may still arrive by \(t_i+\delta\).
The monitor therefore evaluates each formula under the three-valued semantics of runtime verification~\cite{bauer2011}, in its finite-trace form~\cite{kallwies2023anticipatory}: on an open trace, a formula is true if every finite timed trace extending it, including the trace as it stands, satisfies the formula, false if every such extension violates it, and \emph{inconclusive} otherwise.
The pending obligation is thus inconclusive until the trace extends beyond \(t_i + \delta\), since an extension may still supply the witness in time; past that instant no extension can meet the deadline, and the formula is false. Operationally, the monitor reports inconclusive obligations as not-yet-violated, so that no violation is announced before the response is due.

For formulas under \(\mathbf{G}\), the verdict can become true only once the recording ends, since a further observation may falsify their bodies. 
From then on, the only extension is the trace itself, and the three-valued semantics coincides with \(\models\).


When a formula is violated, the monitor also returns the position responsible and the propositions holding there. For globally scoped formulas, this is the earliest failing position; for precedence formulas, the action position with no prior authorization.
This evidence links each reported violation to the observed facts underlying it, making the verdict explainable.

Verdicts are relative to the captured trace. A reported missing authorization may indicate either that no authorization was issued or that it was issued before the trace begins. Distinguishing between these cases requires observations outside the available trace.

\begin{example}
\label{ex:verdict}
The formula of Example~\ref{ex:readbackformula} is evaluated on the trace of Example~\ref{ex:timedtrace}. 
At position~1, the instruction proposition holds, so the implication requires a matching readback within thirty seconds. The readback occurs at \(t_2=8\), within the bound from \(t_1=0\). 
At positions~2 and~3 the instruction proposition is absent, so the implication is true there. Since the outer \(\mathbf{G}\) ranges over all positions, the formula is satisfied. 

If the same trace were still open just after the instruction, with no readback observed yet, the formula would be inconclusive, reported as not-yet-violated, until the trace extended beyond \(t_1+30\); only then would a violation be reported.
\end{example}


\section{Verification Framework}\label{sec:framework}

\begin{figure*}[t]
\centering
\includegraphics[width=\textwidth]{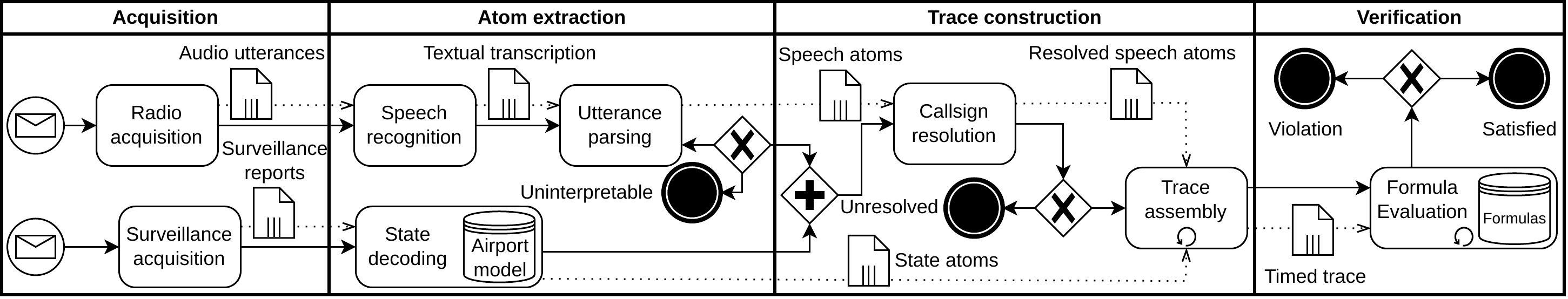}
\caption{Framework architecture and workflow.}
\label{fig:arch}
\end{figure*}

The model in ~\S\ref{sec:formal-model} assumes that atoms and a timed trace are given. 
The architecture in Fig.~\ref{fig:arch} adds four stages: \emph{acquisition} collects heterogeneous operational data, \emph{atom extraction} converts them into the formal vocabulary of ~\S\ref{sec:atomvocabulary}, \emph{trace construction} integrates the atoms into a unified timed trace, and verification evaluates the temporal formulas over it. 
The figure shows offline verification over a completed recording. In the online configuration, the same stages incrementally process a growing trace.

\subsection{Acquisition and atom extraction}

The first two stages transform radio and surveillance signals into ground atoms of \(\mathcal{AP}\) (see ~\S\ref{sec:atomvocabulary}).

Acquisition receives the two raw input signals, each on its own timeline. Radio acquisition captures the voice exchanged between controllers and pilots and outputs the \emph{audio utterances}. Surveillance acquisition collects the aircraft-state messages broadcast through \gls{adsb} and outputs the stream of \emph{surveillance reports}. 
The acquisition stage is modular and can incorporate additional state streams, such as primary radar or aircraft telemetry.

Atom extraction turns the acquired signals into ground atoms: speech atoms are derived from the audio utterances, and state atoms from the surveillance reports.
Speech recognition transcribes each utterance to text; it is a replaceable front end~\cite{aerospace10050490,helmke2022readback}. 
Utterance parsing uses a language model to map each transcription to the speech-atom schema vocabulary. Unmatched transcriptions are discarded as \emph{uninterpretable}.

State decoding outputs two classes of \emph{state atoms}. 
\emph{Air-state atoms} record aircraft level, position, and motion from surveillance reports. 
\emph{Surface-state atoms} relate reported positions to the airport layout to determine runway and taxiway occupancy.

\subsection{Trace construction}

Trace construction converts speech and state atoms into the timed trace \(\pi\) introduced in ~\S\ref{sec:timedtraces}. 
Callsign resolution associates each speech atom with an aircraft and normalizes transcribed or abbreviated call signs against surveillance identities~\cite{aerospace10050490}. 
An unconfirmed call sign is retained only when used consistently by both controller and pilot. 
Unresolved speech atoms are discarded. 
Conversely, state atoms need no resolution: surveillance already tags them with the aircraft.

Trace assembly orders the resolved speech atoms and the state atoms by the time at which the corresponding events occurred, so speech-recognition and decoding delays do not affect the timestamps used by the temporal constraints. 
The trace is incrementally extended with new and derived atoms of \(\mathcal{AP}\): readbacks are matched with the instructions they acknowledge, and clearances are propagated until the action is completed or revoked. Offline, the trace closes when the recording ends. Online, it remains open.

\subsection{Verification}

Formula evaluation receives the timed trace \(\pi\) and the temporal formulas of \(\Phi\) (~\S\ref{sec:temporalpatterns}).
For each family, the monitor selects the formulas whose propositions mention the aircraft and values occurring in \(\pi\), and evaluates them according to ~\S\ref{sec:evalsemantics}.

A formula is evaluated over the timed trace by structural recursion, with the metric operator comparing timestamps against its deadline.
The monitor directly implements the satisfaction relation \(\models\), as in monitors for metric parametric specifications~\cite{basin2015monpoly}. 
Since the outermost operator ranges over all positions and each temporal operator scans at most a suffix of the trace, restricted to its temporal window when the operator is metric, evaluating a formula has worst-case quadratic complexity in the trace length.

In the online configuration, evaluation is also performed between observations by extending the trace with a virtual empty observation at the current instant. 
This allows elapsed deadlines to be detected without waiting for a subsequent event. 
A violation is reported only if a formula is false under both \(\models\) and a second evaluation in which unexpired metric obligations are treated as satisfied.
Formulas for which the two evaluations disagree remain inconclusive.

Verification produces one verdict per formula, two-valued once the recording ends. 
For an open trace, formulas the observations do not yet decide carry the inconclusive verdict; obligations whose deadlines have not yet expired are reported as not-yet-violated.
Any formula that evaluates to false leads to the \emph{violation} outcome; consistent with the explainability mechanism of ~\S\ref{sec:evalsemantics}, each violation identifies the trace position responsible and the propositions holding there.
If no formula is violated, the outcome is \emph{satisfied}.

\section{Evaluation}\label{sec:evaluation}


\subsection{Implementation and experimental setup}
We implement the architecture of Fig.~\ref{fig:arch} as follows.
The acquisition stage can rely on standard radio and \gls{adsb} receiver solutions~\cite{rtlsdr_airband,dump1090antirez}.
In our evaluation, we bypass live acquisition and provide recorded inputs directly (see below).
Speech recognition uses faster-whisper~\cite{fasterwhisper}, a CTranslate2 reimplementation of Whisper~\cite{radford2023whisper}, with a model fine-tuned for air traffic control speech~\cite{tol2024atcwhisper}. The utterance parser is an \gls{llm} performing structured extraction. Inference uses llama.cpp~\cite{llamacpp}, which constrains the output to a JSON-schema grammar; decoding is greedy (temperature~0) with a fixed seed.

State decoding and the airport model are implemented in Python. The airport layout is derived from OpenStreetMap aeroway data~\cite{openstreetmap} and processed with the Shapely geometry library~\cite{shapely}. Callsign resolution, trace assembly, and formula evaluation are implemented as bespoke Python modules. 

For utterance parsing, the online configuration uses \texttt{qwen2.5-7b}~\cite{qwen25} for low latency, while offline analysis uses \texttt{qwen3.8-27b}~\cite{qwen38}. 
We evaluate both thinking and no-thinking modes where available.
The former generates intermediate reasoning and the latter answers directly.

All experiments run locally on a single workstation (NVIDIA RTX~4090, AMD Ryzen~9 7900, 64\,GB RAM, Fedora Linux 44). 
\gls{llm} and \gls{asr} inference run on the GPU, while the remaining framework components run on the CPU.



\subsection{Datasets and vocabulary coverage}\label{sec:datasets}
We feed the pipeline with two public corpora of recorded traffic, ATCO2~\cite{zuluaga2024atco2} and
TartanAviation~\cite{patrikar2025tartanaviation}.
To our knowledge, they are the only publicly available corpora that pair controller--pilot communications with recoverable aircraft state, reflecting the limited availability of \gls{atc} recordings due to legal and regulatory constraints~\cite{zuluaga2024atco2}.

ATCO2 contains communications without aircraft state.
Its openly released one-hour test set provides manually verified transcriptions from seven airports across Europe and Australia, covering tower, ground, approach, and radar operations.
We retrieve the corresponding decoded \gls{adsb} state vectors from the OpenSky Network~\cite{schafer2014opensky}, for the relevant regions and recording intervals, and align them with the radio communications.
TartanAviation provides synchronized radio communications and \gls{adsb} trajectories from a towered airport (Allegheny County Airport, KAGC~\cite{skyvector-kagc}) and a non-towered airport (Pittsburgh--Butler Airport, KBTP~\cite{skyvector-kbtp}). Since the monitored formulas require controller-issued instructions, we restrict the analysis to KAGC.
We form the ground-truth set by annotating in full the busiest hour of each of three days of 2022 KAGC traffic, blind to the monitor's output and using both audio and surveillance.
The set contains every procedural violation present in these three hours: 12 violations, mainly involving readbacks.
ATCO2 transcriptions are passed directly to the parser, whereas TartanAviation audio, for which no manual transcriptions are available, is transcribed by the speech-recognition front end.

We measure vocabulary coverage as the fraction of atom schemas exercised by the corpora, counting a schema when at least one instance appears in the traces. 
It is computed over routine procedures only, since exceptional events such as emergencies, runway incursions, and collision scenarios are absent from the corpora. 
Across both corpora, 37.6\% of the routine schemas are exercised.

\subsection{Evaluation protocol}
The evaluation assesses implementation correctness on synthetic situations and agreement with expert judgment on real traffic. 
We additionally use the case studies (~\S\ref{sec:casestudies}) as qualitative validation.

\noindent
\textbf{Implementation correctness}.
We test the monitor implementation on synthetic situations whose expected verdicts are known by construction. 
Safety-critical configurations are rare in recorded traffic and cannot be deliberately induced in live traffic; we therefore build each situation from a real corpus event satisfying the trigger of a monitored formula, and apply a controlled perturbation like removal of the required response, a deadline exceeded by half a second, or action without the required clearance.
The advisory families are exercised separately through the \"Uberlingen reconstruction of ~\S\ref{sec:accidents}.

\noindent
\textbf{Agreement with expert judgment}.
We evaluate the monitor on real traffic against the ground-truth set of ~\S\ref{sec:datasets}.
Verdicts are scored per event: multiple alerts generated for the same situation across successive surveillance samples count as a single detection.


\subsection{Performance}
\label{sec:eval-performance}

The 255 events satisfying formula triggers yield 1,495 synthetic situations, 664 compliant and 831 violating, exercising 50 forms across five of the eight families of Table~\ref{tab:families}. The monitor returns the expected verdict in every case.
Table~\ref{tab:parser-comparison} reports precision, recall, and F1 of the pipeline of Fig.~\ref{fig:arch} when using the \texttt{qwen3.8-27b}, \texttt{gemma4-26b-a4b}~\cite{gemma4}, and \texttt{qwen2.5-7b} as parsers, scored against the three-hour ground-truth set. Precision is the fraction of reported violations that are correct, recall is the fraction of true violations that are reported, and F1 is their harmonic mean. The no-thinking \texttt{qwen3.8-27b} achieves the highest F1 and recall.

\begin{table}[H]
\centering
\caption{Performance by parser model}
\label{tab:parser-comparison}
\begin{tabularx}{\linewidth}{Xccc}
\hline
\textbf{Parser} & \textbf{Precision} & \textbf{Recall} & \textbf{F1} \\
\hline
\textbf{\texttt{qwen3.8-27b}, no-thinking (NT)} & \textbf{0.79} & \textbf{0.92} & \textbf{0.85} \\
\texttt{qwen3.8-27b}, thinking (T)     & 0.78 & 0.58 & 0.67 \\
\texttt{gemma4-26b-a4b}, no-thinking (NT)     & 0.40 & 0.33 & 0.36 \\
\texttt{gemma4-26b-a4b}, thinking (T)        & 0.67 & 0.33 & 0.44 \\
\texttt{qwen2.5-7b}, no-thinking         & 0.57 & 0.67 & 0.62 \\
\hline
\end{tabularx}
\end{table}

Table~\ref{tab:cost} summarizes the computational cost of each stage using the minimum, first quartile ($\textbf{Q}_1$), median, third quartile ($\textbf{Q}_3$), and maximum over the measured samples. Each stage is timed separately, without concurrent execution of the other stages. Here, \emph{Large LLM} denotes the no-thinking \texttt{qwen3.8-27b} and \emph{Small LLM} the \texttt{qwen2.5-7b} of Table~\ref{tab:parser-comparison}. Speech recognition is reported as inverse real-time factor (RTFx), defined as $t_{\text{audio}}/t_{\text{processing}}$~\cite{srivastav2025openasr}.
Speech recognition consistently operates faster than real time, with limited variability across recordings. Parsing accounts for the largest share of the computational cost, whereas formula evaluation contributes a smaller share despite its wider distribution.

\begin{table}[H]\centering
\caption{Per-stage processing cost.}\label{tab:cost}
\footnotesize
\setlength{\tabcolsep}{1pt}
\begin{tabularx}{\linewidth}{@{}Xcccccl@{}}
\hline
\textbf{Stage} & \textbf{Min} & \textbf{$\textbf{Q}_1$} & \textbf{Median} & \textbf{$\textbf{Q}_3$} & \textbf{Max} & \textbf{Unit} \\
\hline
Speech recognition & 32.1 & 41.4 & 45.1 & 47.5 & 48.7 & RTFx \\
Atom parse (Small LLM) & 0.266 & 0.282 & 0.296 & 0.339 & 1.46 & s/utt. \\
Atom parse (Large LLM) & 1.16 & 1.34 & 1.58 & 1.75 & 4.94 & s/utt. \\
Formula evaluation & 0.000003 & 0.037 & 0.119 & 0.245 & 2.54 & s/event \\
\hline
\end{tabularx}
\end{table}

\begin{figure}[H]\centering
\begin{tikzpicture}
\begin{axis}[width=\columnwidth,height=0.40\columnwidth,xmode=log,
  log ticks with fixed point,
  xlabel={per-utterance parse latency (ms, log scale)},
  ylabel={CDF},ymin=0,ymax=1.02,
  grid=both,grid style={gray!25},tick label style={font=\footnotesize},
  label style={font=\footnotesize},
  /pgfplots/legend image code/.code={\draw[#1] plot coordinates {(0cm,0cm) (0.18cm,0cm)};},
  legend style={font=\tiny,at={(1.00,0.00)},anchor=south east,
    draw=none,fill=white,fill opacity=0.6,text opacity=1,
    inner sep=1pt,nodes={inner sep=1pt},row sep=-2.5pt},
  legend cell align=left]
\addplot[const plot,solid,qwengreen,thick] coordinates {(266,0.0000) (266,0.0556) (272.9,0.1111) (274.3,0.1667) (277.9,0.2222) (281.1,0.2778) (283.7,0.3333) (290.3,0.3889) (290.4,0.4444) (291.9,0.5000) (301,0.5556) (310.1,0.6111) (319.4,0.6667) (329.1,0.7222) (342.4,0.7778) (352.5,0.8333) (375.1,0.8889) (434.2,0.9444) (1463,1.0000)};
\addlegendentry{qwen2.5-7b}
\addplot[const plot,solid,qwenblue,thick] coordinates {(1161,0.0000) (1161,0.0556) (1172,0.1111) (1247,0.1667) (1267,0.2222) (1316,0.2778) (1416,0.3333) (1439,0.3889) (1443,0.4444) (1580,0.5000) (1586,0.5556) (1590,0.6111) (1592,0.6667) (1750,0.7222) (1755,0.7778) (1899,0.8333) (1982,0.8889) (4643,0.9444) (4940,1.0000)};
\addlegendentry{qwen3.8-27b NT}
\addplot[const plot,solid,gemmared,thick] coordinates {(389.6,0.0000) (389.6,0.0556) (424.2,0.1111) (430.5,0.1667) (430.7,0.2222) (438.2,0.2778) (445.6,0.3333) (452.6,0.3889) (453.2,0.4444) (456.5,0.5000) (462.7,0.5556) (513.5,0.6111) (519.8,0.6667) (526.2,0.7222) (532.1,0.7778) (532.4,0.8333) (596.6,0.8889) (1720,0.9444) (1876,1.0000)};
\addlegendentry{gemma4-26b NT}
\addplot[const plot,dashed,qwenblue,thick] coordinates {(10470,0.0000) (10470,0.1000) (14990,0.2000) (19880,0.3000) (27290,0.4000) (36560,0.5000) (39650,0.6000) (46040,0.7000) (51510,0.8000) (55100,0.9000) (83050,1.0000)};
\addlegendentry{qwen3.8-27b T}
\addplot[const plot,dashed,gemmared,thick] coordinates {(308.6,0.0000) (308.6,0.0667) (2270,0.1333) (2311,0.2000) (2438,0.2667) (2588,0.3333) (3104,0.4000) (3148,0.4667) (3950,0.5333) (7296,0.6000) (13920,0.6667) (58560,0.7333) (70040,0.8000) (82790,0.8667) (92880,0.9333) (134400,1.0000)};
\addlegendentry{gemma4-26b T}
\end{axis}
\end{tikzpicture}
\caption{Per-utterance parse latency.}\label{fig:parse-cdf}
\end{figure}
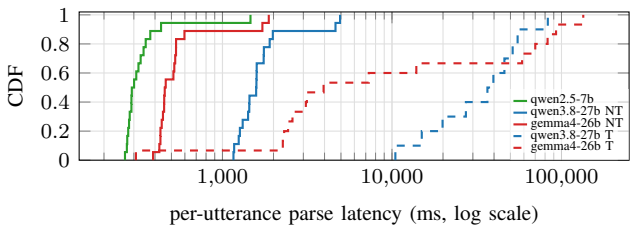

Fig.~\ref{fig:parse-cdf} shows the per-utterance parse latency CDF. The no-thinking distributions are narrow, with medians of about $0.3$\,s (\texttt{qwen2.5-7b}), $0.46$\,s (\texttt{gemma4-26b-a4b}), and $1.6$\,s (\texttt{qwen3.8-27b}). The thinking distributions have medians of about $3.9$\,s (\texttt{gemma4-26b-a4b}) and $38$\,s (\texttt{qwen3.8-27b}), with tails up to $134$\,s and $83$\,s, respectively.

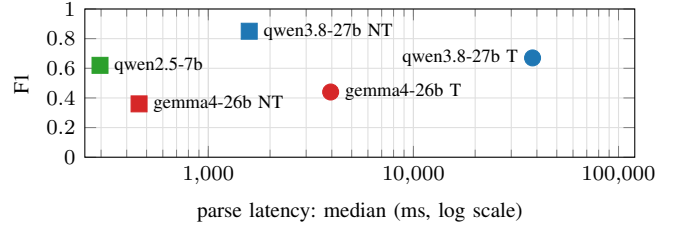
\begin{figure}[H]\centering
\begin{tikzpicture}
\begin{axis}[width=\columnwidth,height=0.40\columnwidth,xmode=log,
  log ticks with fixed point,
  xlabel={parse latency: median (ms, log scale)},
  ylabel={F1},ymin=0,ymax=1,xmin=250,xmax=120000,
  grid=both,grid style={gray!25},tick label style={font=\footnotesize},label style={font=\footnotesize}]
\addplot[only marks,mark=square*,qwenblue,mark size=3pt] coordinates {(1583,0.85)};
\addplot[only marks,mark=square*,gemmared,mark size=3pt] coordinates {(459.6,0.36)};
\addplot[only marks,mark=*,qwenblue,mark size=3pt] coordinates {(38100,0.67)};
\addplot[only marks,mark=*,gemmared,mark size=3pt] coordinates {(3950,0.44)};
\addplot[only marks,mark=square*,qwengreen,mark size=3pt] coordinates {(296.5,0.62)};
\node[font=\scriptsize,anchor=west] at (axis cs:1583,0.85) {\;qwen3.8-27b NT};
\node[font=\scriptsize,anchor=west] at (axis cs:459.6,0.36) {\;gemma4-26b NT};
\node[font=\scriptsize,anchor=east] at (axis cs:38100,0.67) {qwen3.8-27b T\;};
\node[font=\scriptsize,anchor=west] at (axis cs:3950,0.44) {\;gemma4-26b T};
\node[font=\scriptsize,anchor=west] at (axis cs:296.5,0.62) {\;qwen2.5-7b};
\end{axis}
\end{tikzpicture}
\caption{Accuracy versus parse latency.}\label{fig:time-f1}
\end{figure}

Fig.~\ref{fig:time-f1} relates accuracy to latency.
The no-thinking \texttt{qwen3.8-27b} sits at the highest F1 at moderate latency, the \texttt{qwen2.5-7b} at the lowest latency with reduced F1, and the thinking configurations at much higher latency and lower F1. The 7B therefore suits the online monitor, while the no-thinking \texttt{qwen3.8-27b} serves as the offline reference parser.

\subsection{Documented accidents}
\label{sec:accidents}

The two accidents introduced in ~\S\ref{sec:casestudies} are replayed through the monitor, with their radio and surveillance state reconstructed from the official investigation reports.
The cases lie outside the corpora from ~\S\ref{sec:datasets} and as such carry no accuracy metrics.
They test the monitor on rare safety-critical situations, with the official reports as reference: in each case we check whether the firings agree, in content and timing, with the events the investigators documented, and whether the detection instants leave a margin usable for a real-time alert.

We report detection instants on the wall clock of the investigation reports.
The monitor itself runs on session-relative time and places each violation at the instant its formula is decided over the reconstructed trace, once the deadline following the triggering observation has elapsed.

\begin{figure}[H]\centering
\resizebox{\columnwidth}{!}{%
\begin{tikzpicture}[font=\scriptsize,>={Latex[length=1.2mm]},
  ev/.style={align=center,inner sep=1.5pt},
  fire/.style={align=center,draw,rounded corners=2pt,fill=black!8,inner sep=2.5pt},
  lane/.style={font=\scriptsize\itshape,text=gray},
  lead/.style={thick,->}]
\draw[-{Latex[length=2mm]}] (0.05,0) -- (8.6,0);
\node[lane,rotate=90] at (-0.30,0.85) {Monitor};
\node[lane,rotate=90] at (-0.30,-0.85) {BFU};
\node[fire,anchor=south] (f1) at (2.30,0.5) {21:35:01\\advisory-direction\\resolution-advisory};
\node[fire,anchor=south] (f2) at (5.4,0.5) {21:35:03\\advisory-priority};
\draw[lead] (2.30,0) -- (f1.south);
\draw[lead] (2.66,0) -- (2.66,0.22) -| (f2.south);
\node[ev,anchor=north] (e1) at (1.40,-0.5) {21:34:56\\Tu-154 climb RA,\\descends per ATC};
\node[ev,anchor=north] (e2) at (5.4,-0.5) {21:35:03\\ATC orders descend\\(against the RA)};
\draw[lead] (1.40,0) -- (e1.north);
\draw[lead] (2.66,0) -- (2.66,-0.22) -| (e2.north);
\draw (7.79,0.10)--(7.97,-0.10); \draw (7.79,-0.10)--(7.97,0.10);
\node[ev,anchor=north] at (7.88,-0.5) {21:35:32\\collision};
\end{tikzpicture}}
\caption{\"Uberlingen: monitor detections and BFU-documented events (UTC).}
\label{fig:uberlingen-timeline}
\end{figure}
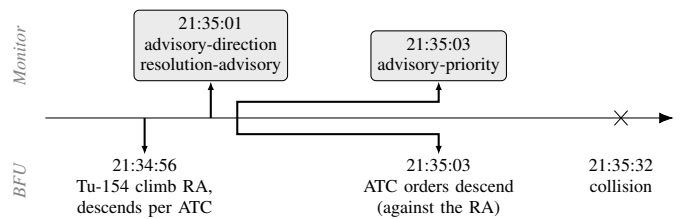

\noindent
\textbf{\"Uberlingen}.
On the reconstructed radio, avionics, and surveillance state, the monitor flags the same sequence the BFU documented~\cite{bfu2004ueberlingen}. At 21:34:56 the Tupolev's collision-avoidance system commands a climb while surveillance shows it descending under the controller's standing instruction; \emph{advisory-direction} and \emph{resolution-advisory} are confirmed at 21:35:01, once the reaction window elapses, and the monitor also flags that the advisory is never reported to the controller, as \gls{icao} requires of a crew deviating in response to one~\cite[\S15.7.3.3,~Note]{InternationalCivilAviationOrganization2016Procedures}. At 21:35:03 a second descent is issued while the advisory is active, firing \emph{advisory-priority}.
Each of these firings lies in the avionics and surveillance state, not on the radio, and those at 21:35:01 and 21:35:03 precede the 21:35:32 collision by thirty-one and twenty-nine seconds, respectively (Fig.~\ref{fig:uberlingen-timeline}).
Thus the monitor reproduces the precedence reasoning the BFU documented.

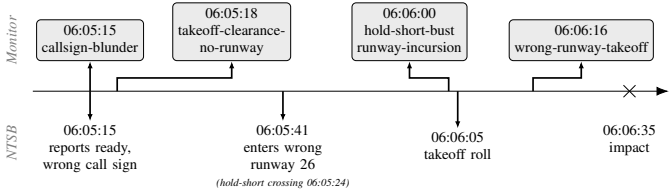
\begin{figure}[H]\centering
\resizebox{\columnwidth}{!}{%
\begin{tikzpicture}[font=\scriptsize,>={Latex[length=1.2mm]},
  ev/.style={align=center,inner sep=1.5pt},
  fire/.style={align=center,draw,rounded corners=2pt,fill=black!8,inner sep=2.5pt},
  lane/.style={font=\scriptsize\itshape,text=gray},
  lead/.style={thick,->}]
\draw[-{Latex[length=2mm]}] (0.05,0) -- (10.6,0);
\node[lane,rotate=90] at (-0.30,0.85) {Monitor};
\node[lane,rotate=90] at (-0.30,-0.85) {NTSB};
\node[fire,anchor=south] (g1) at (1.0,0.5) {06:05:15\\callsign-blunder};
\node[fire,anchor=south] (g1b) at (3.35,0.5) {06:05:18\\takeoff-clearance-\\no-runway};
\node[fire,anchor=south] (g2) at (6.3,0.5) {06:06:00\\hold-short-bust\\runway-incursion};
\node[fire,anchor=south] (g3) at (9.15,0.5) {06:06:16\\wrong-runway-takeoff};
\draw[lead] (1.0,0) -- (g1.south);
\draw[lead] (1.45,0) -- (1.45,0.22) -| (g1b.south);
\draw[lead] (6.95,0) -- (6.95,0.22) -| (g2.south);
\draw[lead] (8.35,0) -- (8.35,0.22) -| (g3.south);
\node[ev,anchor=north] (h1) at (1.0,-0.5) {06:05:15\\reports ready,\\wrong call sign};
\node[ev,anchor=north] (h2) at (4.2,-0.5) {06:05:41\\enters wrong\\runway 26\\{\tiny\itshape (hold-short crossing 06:05:24)}};
\node[ev,anchor=north] (h3) at (7.1,-0.62) {06:06:05\\takeoff roll};
\draw[lead] (1.0,0) -- (h1.north);
\draw[lead] (4.2,0) -- (h2.north);
\draw[lead] (7.1,0) -- (7.1,-0.30) -| (h3.north);
\draw (9.85,0.10)--(10.05,-0.10); \draw (9.85,-0.10)--(10.05,0.10);
\node[ev,anchor=north] at (9.95,-0.5) {06:06:35\\impact};
\end{tikzpicture}}
\caption{Comair: monitor detections and NTSB-documented events (EDT).}
\label{fig:comair-timeline}
\end{figure}

\noindent
\textbf{Comair Flight 5191 (Comair)}.
On the reconstructed radio and position track, the monitor flags the sequence the NTSB documented~\cite{ntsb2007comair5191}. At 06:05:15 the first officer reports ready with the wrong call sign, ``Comair one twenty one'' for one ninety one, firing the \emph{callsign-blunder} phraseology check. The authorization on record is runway~22, named in the taxi clearance; the takeoff clearance names no runway. \emph{takeoff-clearance-no-runway} fires at 06:05:18, six seconds before the hold-short crossing. The formula encodes the current requirement of a runway designator in the takeoff clearance~\cite[\S12.3.4.11]{InternationalCivilAviationOrganization2016Procedures}; this requirement was not in force at the time of the accident~\cite{ntsb2007comair5191}. As the track crosses onto the surface the monitor registers runway entry at 06:06:00 and fires \emph{runway-incursion} and its precursor check \emph{hold-short-bust}. This multi-source detection compares the runway authorized on the radio with the runway the aircraft actually occupies.
Both firings are decided once the reconstruction confirms surface entry at 06:06:00, and so trail the documented hold-short crossing at 06:05:24; that earlier instant is not resolved in this reconstruction. \emph{wrong-runway-takeoff}, a specialization of the \emph{runway-incursion} family, fires at 06:06:16, as ground speed enters the takeoff-roll regime, nineteen seconds before the 06:06:35 impact in Fig.~\ref{fig:comair-timeline}.
Thus the monitor identifies the wrong-runway takeoff while the roll is still in progress.

\section{Discussion}\label{sec:discussion}
\noindent
\textbf{Performance figures}.
The F1 of 0.85 on real traffic reflects the whole chain that recovers observations from speech: the conditions of the radio channel, the speech-recognition model, and the parser.
Hold the formulas and the recordings fixed, and the score moves with the parser alone (Table~\ref{tab:parser-comparison}); each figure also carries the quality of the transcriptions it reads. Advances in speech recognition and language models therefore carry straight over to monitoring, with the formulas and their semantics untouched.
The computational requirements of the system are compatible with deployment on standard workstation hardware. The full pipeline executes on a single machine, with parsing as the dominant computational cost. The lightweight parser, \texttt{qwen2.5-7b}, operates with sub-second latency and could support real-time monitoring; the slower but more accurate parser, \texttt{qwen3.8-27b}, suits offline analysis of recorded operational data.

\noindent
\textbf{Operational implications}.
The evaluation shows that the four monitoring requirements can be addressed jointly: operational traffic exercises grounding (\reqGrounding), the accident reconstructions combine radio with surveillance or onboard observations (\reqIntegration), and both require temporal ordering and deadlines (\reqDeadlines); every reported violation remains linked to the failed obligation and its supporting observations (\reqVerdicts).
Beyond this requirement coverage, the results support the three uses outlined in ~\S\ref{sec:intro}. 
For real-time assistance, early detection can support risk mitigation by identifying deviations while the operational state is still evolving, as in the reconstructed accidents. 
Operationally, such early detection provides a basis for risk mitigation. 
While the aircraft's dynamics and controllability remain beyond the monitor's purview, identifying a deviation while the operational state is still evolving creates a window for potential intervention, particularly when a loss of situational awareness is a primary trigger or a contributing factor.
For post-operation review, each violation is linked to the failed obligation and supporting observations, providing an immediate procedural account before the completion of a formal investigation. 
The same evidence can support training by providing explicit feedback on procedural deviations in recorded sessions, where the more accurate offline parser can be used without real-time latency constraints.
Nevertheless, operational deployment requires human-in-the-loop evaluation to study how the tool integrates with existing ATC workflows and systems, and how its alerts, including false positives, should be presented and calibrated under controller workload.

\section{Related Work}\label{sec:related}
%

Existing work addresses complementary parts of spoken ATC procedure monitoring. 
Table~\ref{tab:related} summarizes the closest representative approaches with respect to the four requirements introduced in ~\S\ref{sec:intro}.
A requirement is met (\hbF{}) when the approach represents it and uses it in the monitoring decision, partially met (\hbH) when it is supported only in a restricted or indirect form, and not addressed (\hbE{}) when absent from the method's inputs, model, or verdict.

\begin{table}[!b]
\centering
\caption{Requirement coverage: related work vs. this work.}
\label{tab:related}
\footnotesize
\setlength{\tabcolsep}{3pt}
\begin{tabularx}{\linewidth}{@{}Xccccccc@{}}
\hline
 & \cite{reynolds2002conformance} & \cite{helmke2022readback} & \cite{lin2020atc} & \cite{lima2023explainable} & \cite{maierhofer2020interstate,krasowski2021colregs} & \cite{lall2025promptcheck} & \textbf{This work} \\
\hline
\reqGrounding{} grounding & \hbE & \hbF & \hbF & \hbE & \hbE & \hbH & \hbF \\
\reqIntegration{} integration & \hbE & \hbH & \hbF & \hbE & \hbE & \hbE & \hbF \\
\reqDeadlines{} timeliness & \hbH & \hbH & \hbH & \hbF & \hbF & \hbE & \hbF \\
\reqVerdicts{} verdicts & \hbE & \hbH & \hbH & \hbH & \hbH & \hbH & \hbF \\
\hline
\multicolumn{8}{@{}l@{}}{\scriptsize \hbF{}~met \quad \hbH{}~partially met \quad \hbE{}~not addressed}\\
\end{tabularx}
\end{table}

On the surveillance side, Reynolds et al.~\cite{reynolds2002conformance} compare observed tracks with trajectories expected from active clearances. Trajectory timing provides a restricted form of \reqDeadlines, while radio exchanges are not an input (\reqGrounding, \reqIntegration), and alerts identify the aircraft but not the breached obligation or supporting observations (\reqVerdicts). Thus, \reqDeadlines\, is partially met, while the other requirements are not addressed.

On the voice side, a common ontology~\cite{helmke2018ontology} structures transcribed controller and pilot utterances by command type, call sign, and values. 
Building on it, Helmke et al.~\cite{helmke2022readback} compare pilot readbacks with controller instructions, linking utterances to the addressed aircraft and instructed values (\reqGrounding). 
Surveillance restricts candidate call signs during recognition but does not enter the compliance check, which stops at the readback rather than extending to the maneuver actually flown (\reqIntegration); timing is limited to a response timeout (\reqDeadlines); and the output flags discrepant values rather than the underlying obligation (\reqVerdicts). 
\reqGrounding{} is met; \reqIntegration, \reqDeadlines, and \reqVerdicts{} are partially met.

Combining voice and surveillance, Lin et al.~\cite{lin2020atc} ground spoken instructions and evaluate them together with the \gls{adsb} track, meeting \reqGrounding{} and \reqIntegration{} in a single compliance decision. 
Their pipeline implements three checks: repetition, conflict detection, and conformance with the instructed maneuver, the latter measured against a trajectory predicted from the instruction. 
Repetition and conformance use fixed validity windows, but instructions do not supersede one another, providing only partial \reqDeadlines. 
Warnings identify the triggered check, recognized text, aircraft, and time, but not the breached obligation, partially addressing \reqVerdicts.


Among \gls{rv} monitors (\S\ref{sec:ltl}), Lima et al.~\cite{lima2023explainable} explain each verdict with a proof at the formula level. 
Their specification language expresses deadlines and ordering, meeting \reqDeadlines. 
\reqVerdicts{} is only partially met, since the explanation remains at the formula level rather than linking the verdict to the underlying operational observations. 
The trace is assumed given as logical facts, so \reqGrounding{} and \reqIntegration{} are not addressed.

In neighboring transportation domains, Maierhofer et al.~\cite{maierhofer2020interstate} formalize interstate driving norms in \gls{mtl} and evaluate them over recorded trajectories, while Krasowski et al.~\cite{krasowski2021colregs} do the same for the International Regulations for Preventing Collisions at Sea (COLREGS). 
Metric operators encode rule time bounds (\reqDeadlines), and violations identify the affected rule, partially addressing \reqVerdicts. 
Because the trajectory is the single observation source, \reqGrounding{} and \reqIntegration{} are not addressed.

Interpretation can also be delegated to a language model. 
Lall et al.~\cite{lall2025promptcheck} prompt models to judge protocol compliance in maritime training transcripts. 
Utterances are linked to checklist items rather than identified entities or values, providing only partial \reqGrounding; the transcript is the sole input, so \reqIntegration{} is not addressed. 
Time selects the excerpts presented to the model without a procedural deadline, so \reqDeadlines{} is not addressed. 
Correctness is assessed against expert annotation; reported items include an instant and quoted span, partially addressing \reqVerdicts.

In summary, existing approaches cover different subsets of the requirements, while our framework addresses them jointly.

\section{Conclusion and Future Work}\label{sec:conclusion}

We presented an executable and explainable framework for monitoring spoken ATC procedures. It grounds communications in the aircraft and values they concern, integrates them with surveillance and onboard observations in a timed trace, and evaluates \gls{icao}-derived obligations with explicit deadlines and precedence. Violations are linked to the failed obligation and supporting observations.

Future work will extend the set of monitored formulas to cover local procedural conventions, validate the framework on broader traffic corpora, and assess its integration into operational \gls{atc}. 
We also plan to evaluate the approach in other transportation domains.



\bibliographystyle{ieeetran}
\bibliography{refs}

\vskip -2\baselineskip plus -1fil

\begin{IEEEbiography}[{\includegraphics[width=1in,height=1.25in,clip,keepaspectratio]{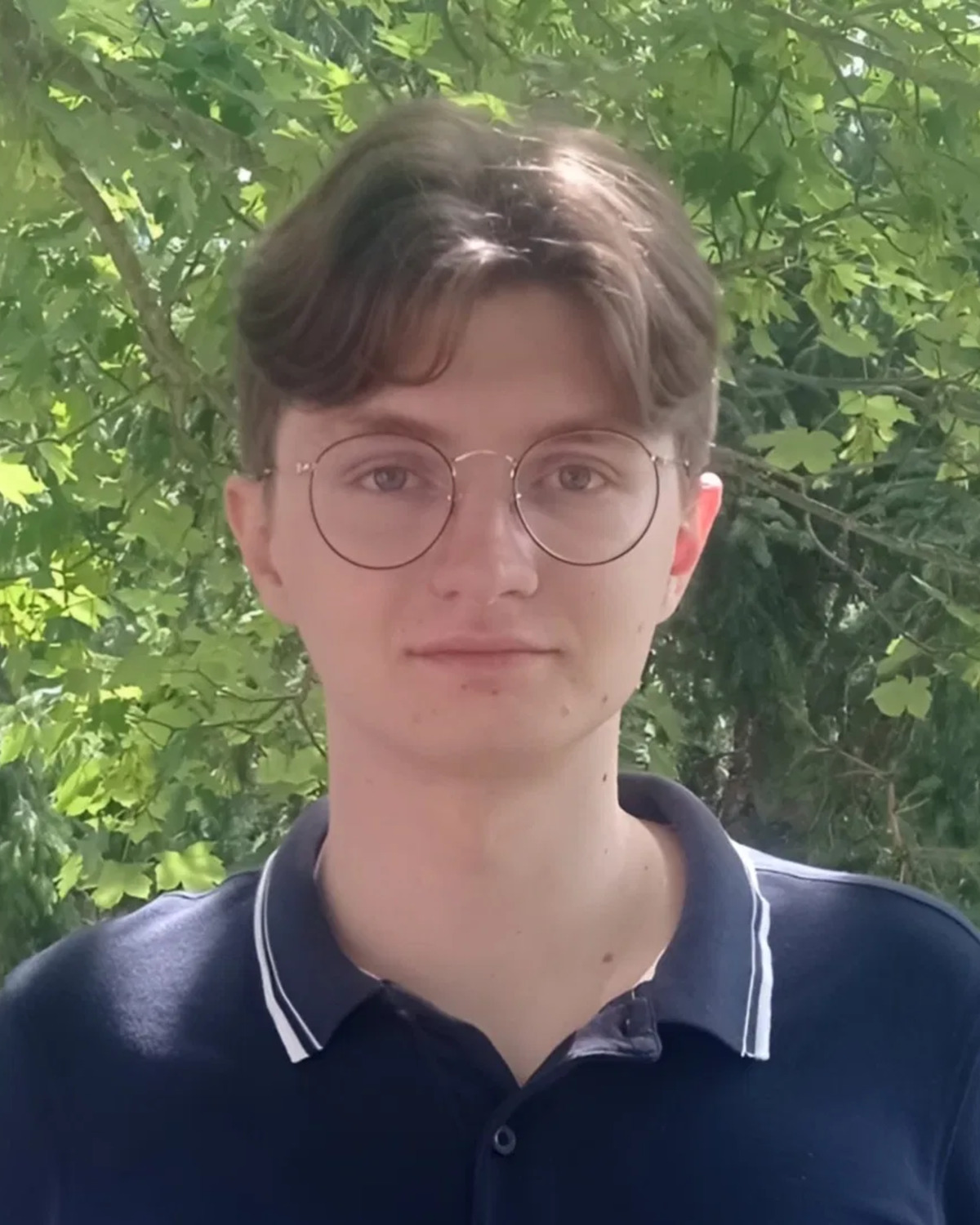}}]{Roberto Luvini}
is an MSc student in Computer Engineering (Software Platforms and Cybersecurity) at the University of Genoa, Italy. His research interests include cybersecurity and air traffic operations. He has been admitted to a PhD programme in Cybersecurity and Artificial Intelligence at the University School of Advanced Defense Studies (CASD), where he will continue his research in this area.
\end{IEEEbiography}

\vskip -2\baselineskip plus -1fil

\begin{IEEEbiography}[{\includegraphics[width=1in,height=1.25in,clip,keepaspectratio]{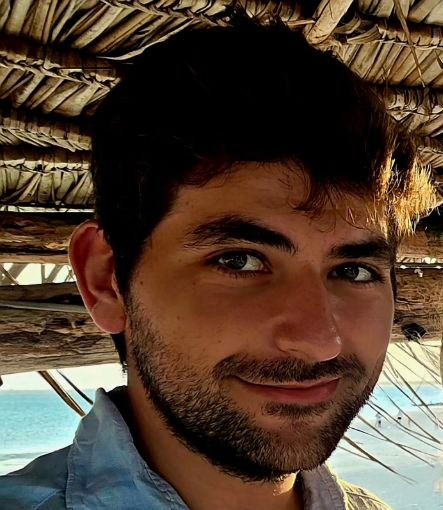}}]{Giacomo Longo} is a Contract Researcher at the University School of Advanced Defense Studies (CASD), Rome, Italy.
He holds a Ph.D. in Artificial Intelligence for Security from Sapienza University of Rome and a Master's degree in Computer Engineering from the University of Genoa.
His research interests include the cybersecurity of cyber-physical and transportation systems (maritime, avionic, and industrial control systems), wireless security, and digital forensics.
\end{IEEEbiography}

\vskip -2\baselineskip plus -1fil

\begin{IEEEbiography}[{\includegraphics[width=1in,height=1.25in,clip,keepaspectratio]{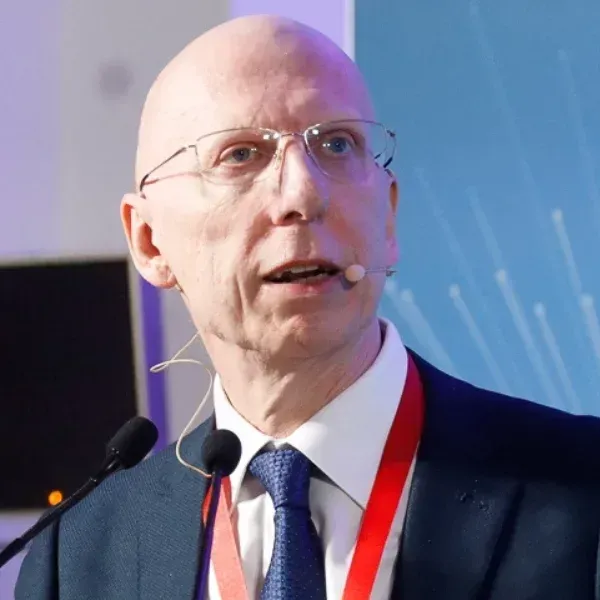}}]{Alessandro Armando} is a Professor of Computer Security at the University of Genoa. His research focuses on the application of automated reasoning techniques to the security of distributed systems. He has coordinated and led research teams in several national and European projects. He formerly served as Director of the CINI National Cybersecurity Laboratory and currently chairs the Scientific Committee of the SERICS Foundation.
\end{IEEEbiography}

\vskip -2\baselineskip plus -1fil

\begin{IEEEbiography}[{\includegraphics[width=1in,height=1.25in,clip,keepaspectratio]{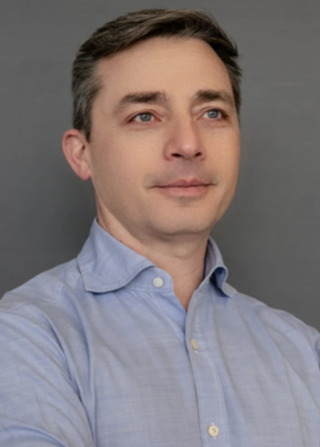}}]{Enrico Russo} received his M.Sc.in Computer Science and Ph.D. in Computer Science and Systems Engineering at the University of Genoa in 2001 and 2021.
He joined the University School of Advanced Defense Studies (CASD), Rome, Italy, as an Assistant Professor in 2026.
His research focuses on the cybersecurity of cyber-physical systems, particularly in the transportation domain, leveraging cyber range capabilities for security testing and assessment.
\end{IEEEbiography}

\end{document}